\documentclass{article}

     \PassOptionsToPackage{numbers, compress}{natbib}
\usepackage{subcaption}
\usepackage{graphicx}
\usepackage{placeins}

\usepackage{multirow}

  \usepackage[dblblindworkshop, final]{neurips_2026}
\workshoptitle{Interpretability for Discovery}

\usepackage[utf8]{inputenc} 
\usepackage[T1]{fontenc}    
\usepackage{hyperref}       
\usepackage{url}            
\usepackage{booktabs}       
\usepackage{amsfonts}       
\usepackage{nicefrac}       
\usepackage{microtype}      
\usepackage{xcolor}         
\usepackage{amsmath}

\title{Higher Order Component Attribution via Learned Surrogates in High Energy Physics}

\author{ 
  \bfseries   Lauri Laatu$^{1}$,
  Aritra Bal$^{2,3}$,
  Benedikt Maier$^{1}$, \\ 
  \bfseries Markus Klute$^{4}$,
  Michael Spannowsky$^{2,3}$
  \\[1.0em]
  \begin{tabular}{c}
    $^{1}$Imperial College London, United Kingdom \\
    $^{2}$Institute for Theoretical Physics (ITP), Karlsruhe Institute of Technology, Germany \\
    $^{3}$Institute for Quantum Materials and Technologies (IQMT), Karlsruhe Institute of Technology, Germany \\
    $^{4}$Institute of Experimental Particle Physics (ETP), Karlsruhe Institute of Technology, Germany \\
  \end{tabular}
}

\begin{document}

\maketitle

\begin{abstract}
Experimental High Energy Physics (HEP) has adopted the transformer architecture for identification of jets, particularly the state-of-the-art Particle Transformer (ParT) which injects explicit information about particle correlations into the attention mechanism. While able to achieve state-of-the-art results in accuracy, interpretability to its effectiveness has been limited. In this paper we propose component attribution methods using Kolmogorov-Arnold Networks (KANs) that directly tie back to the physical descriptors of jets, to probe whether mechanisms such as Multi-Head Attention (MHA), or the ParT architecture, are able to learn these as discriminating features.

\end{abstract}

\section{Introduction}
We begin with a quick look at the physics of jets, a kind of object recorded at particle physics experiments such as the Large Hadron Collider (LHC). These have a complex internal radiation pattern which high-energy physicists study to gain a deeper understanding of the underlying physics governed by the Standard Model (SM) or to search for hints of what lies beyond.


\subsection{Jets at Collider Experiments}
\begin{figure}[htbp]
  \centering
  \begin{subfigure}{0.32\textwidth}
    \centering
    \includegraphics[width=\linewidth]{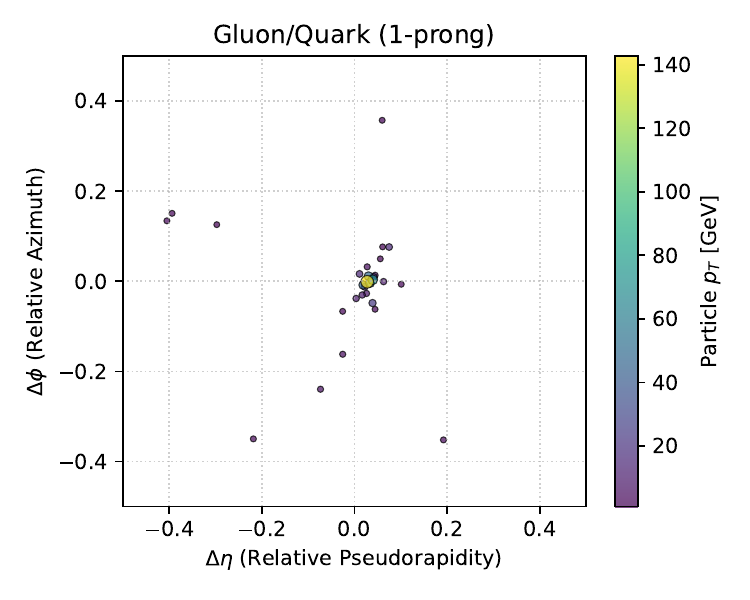}
    \label{fig:sink1}
  \end{subfigure}\hfill
  \begin{subfigure}{0.32\textwidth}
    \centering
    \includegraphics[width=\linewidth]{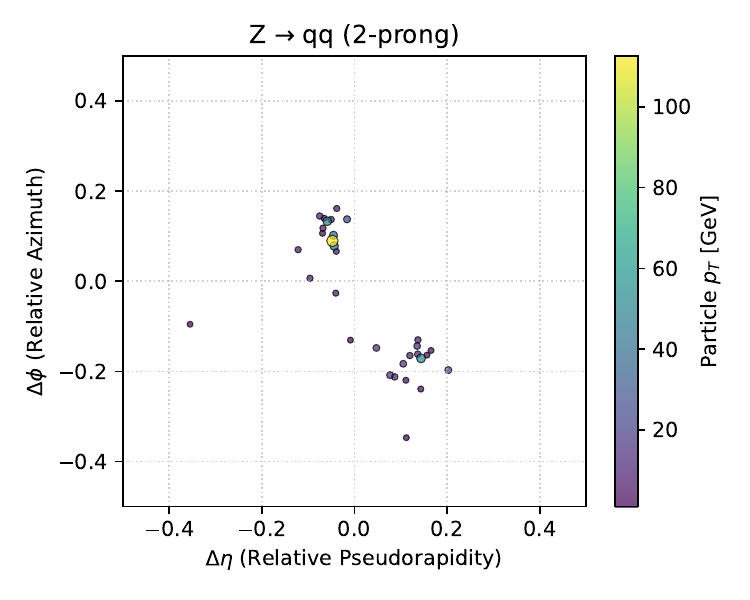}
    \label{fig:sink2}
  \end{subfigure}\hfill
  \begin{subfigure}{0.32\textwidth}
    \centering
    \includegraphics[width=\linewidth]{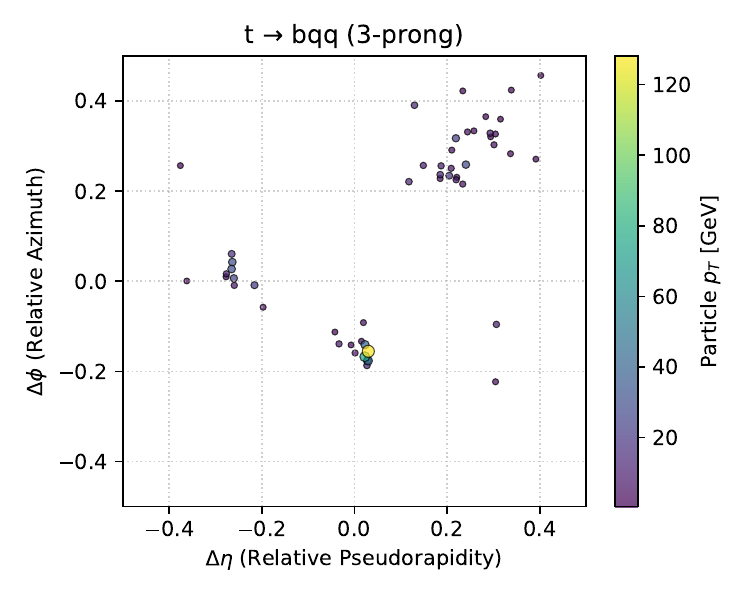}
    \label{fig:sink3}
  \end{subfigure}
  
  \caption{Representative jet constituents in the $(\eta,\phi)$ plane, centered on the jet axis, for one-, two-, and three-pronged jet topologies. From left to right: a light quark- or gluon-initiated jet (one prong), a jet from a $Z \to q\bar{q}'$ decay (two prongs), and a jet from a hadronically decaying top quark, $t \to bq\bar{q}'$ (three prongs). }
  \label{fig:jet_prongs}
\end{figure}

Jets are narrow cone-shaped streams of particles produced when high energy elementary particles such as quarks or gluons are ejected from a collision. Because of color confinement, quarks and gluons cannot exist as free particles. Instead, they undergo hadronization, a process that leads to a shower of hadrons, these being composite particles which may decay further until only stable or long-lived hadrons remain to be detected, as signatures in the detector. When originating from the decay of high transverse-momentum (which we refer to as boosted) massive intermediate particles, the internal substructure of jets exhibits for example, two (such as a $W/Z$ boson) or three (top quark) pronged patterns.  Distributions of the jet constituent particles in the $\eta - \phi$ plane that highlight these patterns are shown in Figure \ref{fig:jet_prongs}. Here $\eta$ and $\phi$ denote the pseudorapidity and azimuthal angle \footnote{The azimuthal angle $\phi$ measures position around the collider beam axis. The pseudorapidity $\eta$ is defined as $\eta \equiv -\ln[\tan(\theta/2)]$, with $\theta$ being the polar angle.} of each constituent, measured relative to the jet axis rather than to the beam axis

Distinguishing these classes of jets against the standard QCD background originating from light quarks or gluons has often, in the past, relied on observables built using the kinematic properties of individual jet constituents, rather than global ones. The energy-energy correlator (EEC), introduced in \cite{Basham:1978bw}, quantifies the correlated energy flow between pairs of final-state particles as a function of their angular separation. Similarly, energy correlation functions (ECFs) \cite{Larkoski:2013eya} extend the pairwise correlation to arbitrary particle multiplicities within a single jet using infrared-and-collinear (IRC) safe combinations of $p_\mathrm{T}$ fractions and pairwise angular separations. An $N$-point energy correlation function (ECF) measures how energy is shared among groups of $N$ particles in a jet, weighting each group by the particles’ energies and their pairwise angular separations. As the definition of Equation \ref{eq:ecf_general} implies, contributions are suppressed if any particle has low energy or a given pair is close in angle. This observable therefore probes whether the jet contains at least $N$ distinct prongs, and ratios built from these serve as a powerful variable for jet flavor tagging.

For a jet $J$ reconstructed at a hadron collider, the general $N$-point energy correlation function is:

\begin{equation} \mathrm{ECF}(N,\beta) = \sum_{i_1 < i_2 < \cdots < i_N \,\in\, J} \left(\prod_{a=1}^{N} z_{i_a}\right) \left(\prod_{(a,b)} \theta_{i_a i_b}\right)^{\beta}, \label{eq:ecf_general}
 \end{equation}

where the sum runs over all distinct $N$-tuples of constituents of $J$; $z_i \equiv p_{T,i}/p_{T,J}$ is the transverse-momentum fraction of constituent $i$ relative to the jet; $\theta_{ij} \equiv \Delta R_{ij} = \sqrt{(\Delta\eta_{ij})^2 + (\Delta\phi_{ij})^2}$ is the angular separation between constituents $i$ and $j$ in the rapidity--azimuth plane. 

The second product runs over all $\binom{N}{2}$ pairs $(a,b)$ drawn from the $N$-tuple; and $\beta > 0$ sets the angular weighting exponent. Building the ECF from $p_T$ ratios and $\Delta R$, rather than from energies and opening angles as is the case at $e^+e^-$ colliders, keeps it invariant under longitudinal boosts along the beam axis. The two- and three-point correlators used in this work are referred to as $e_2^{(\beta)} \equiv \mathrm{ECF}(2,\beta)$ and $e_3^{(\beta)} \equiv \mathrm{ECF}(3,\beta)$ and defined as follows: 

\begin{align} 
e_2^{(\beta)} &= \sum_{i < j \,\in\, J} z_i z_j\, \theta_{ij}^{\beta}, \label{eq:e2} \\
e_3^{(\beta)} &= \sum_{i < j < k \,\in\, J} z_i z_j z_k\, \left(\theta_{ij}\, \theta_{ik}\, \theta_{jk}\right)^{\beta}. 
\label{eq:e3} 
\end{align}

At $\beta = 1$, $e_2^{(1)}$ sums the pairwise product of $p_T$ fractions weighted linearly by the angular separation of each pair, and so measures how widely the jet's energy is spread in angle. $e_3^{(1)}$ extends the same construction to constituent triplets, weighting the combined $p_T$ fraction of each triplet by the product of the three possible pairwise separations. This quantity is most sensitive to the three-pronged angular structure of the type observed in fully hadronic top-quark decays. 

\subsection{Machine Learning for Jet Tagging}
Deep neural network (DNN) architectures are the tool of choice for most jet tagging applications at collider experiments. Large parameter counts, often on the order of millions, ensure they learn the radiation patterns and correlations within a jet. This removes any reliance on a finite set of hand-crafted observables such as the ECFs above. Learning correlations from the raw kinematic data generally translates into higher metrics such as the AUC score, or the background rejection at fixed signal efficiency, across signal models. 

Early DNN-based taggers represented a jet as a fixed-size image, with the particle momenta being deposited in $\eta$--$\phi$ pixels. This representation allowed the usage of Convolutional Neural Network (CNN) architectures \cite{Cogan:2015, deOliveira:2015xxd, Komiske:2016rsd} that had already demonstrated great success in computer vision tasks. But this approach did not consider the variable-multiplicity, point-cloud nature of the jet, and in turn motivated architectures that directly operate on the constituents themselves. For example, permutation-invariant deep sets \cite{Komiske:2018cqr}, and graph neural networks (GNNs) such as ParticleNet \cite{Qu_2020} and DeepJet \cite{Bols:2020bkb}. Transformer architectures, such as the Particle Transformer \cite{Qu:2022mxj} outperform both the convolutional and graph-based taggers above. These form the backbone architecture used in most modern jet tagging applications, and are also the focus of this work.

 The volume of work examining how these networks arrive at classification decisions is not so numerous. Previous work using mechanistic interpretability methods in high-energy physics include~\cite{bonnetguerrini2026findingusinginterpretablelatents,rai2026dissectingjettaggermechanisticinterpretability,gambhir2026interpretinginterpretabilityexplainingexplainability}. A general question remains, can a multi-million-parameter model's decisions be related or traced back to the jet's substructure and the radiation patterns within it. One possible route is to test whether a trained network assigns the largest attention scores or gradient-based importance, to the same particle groupings (pairs or triplets) that contribute the most to the observables defined in Equations \ref{eq:e2} and \ref{eq:e3}.

\section{Methods}

\subsection{Component Attribution and the Need for Higher-Order Evaluation}

Component attribution provides a powerful mechanism to directly read off the “contribution” of every component to a prediction. This effectively abstracts away the complexity of the model’s internal computation. Advanced ablation techniques, such as Component Attribution via Regression (COAR)~\citep{coar}, can accurately predict how model predictions change in response to component-level ablations and assign a precise score to component importance. However, many such methods assume additive linear effects between components. While first-order terms can quantify independent effects, higher-order terms are required to reveal joint interactions that might not be pairwise-irreducible. In the context of jet tagging, it is expected that such terms are necessary to study the internal radiation pattern of high-multiplicity jets. A standard pairwise graph can encode covariances among observables, but it cannot tell whether a triangle of strong pairwise links came from one irreducible three-observable fluctuation or from three unrelated pairwise effects~\cite{bal2026informationgeometryjetsubstructure}.

\subsection{Higher-Order Component Attribution via KAN Surrogates}

We propose higher-order component attribution via Kolmogorov-Arnold Networks (KANs)~\cite{mammeri2025highorder} to better understand the decision mechanism within a DNN-based jet tagger. Our approach attempts to study whether ECFs, more specifically the $e_2^{(1)}$ and $e_3^{(1)}$ correlators, can be related to the neural network's learned outcome, therefore telling us whether large multi-million parameter networks still learn the same physics. We can then determine whether architectures like the Particle Transformer (ParT)~\cite{Qu:2022mxj} learn higher-order terms and irreducible patterns within a jet, or rely on algorithmic shortcuts. By starting from a fully trained target model, we first apply localized Gaussian perturbations to the input particle kinematic vectors. We then train a KAN with cubic b-spline activations as a local surrogate that maps these input perturbations to the target model's output logit shifts. A single cubic spline is continuously differentiable twice, but the off-diagonal interaction terms we use are well-defined. Each edge of the KAN acts on a single variable, so a mixed partial derivative over distinct particles involves only the first derivatives of the input-layer splines, these being continuous everywhere. The higher derivatives come from these deeper splines, evaluated at hidden activations that lie away from the spline knots. This b-spline formulation allows the surrogate to fit the perturbation-response surface without suffering from global oscillatory artifacts such as Runge's phenomenon.


We broadly follow the method set out in~\cite{mammeri2025highorder}: assume a set of kinematic inputs $x \in \mathbb{R}^n$. Extending the local-proxy view, we define the order-$r$ interaction coefficient for index set $S$ (with $\vert{}S\vert{} = r$) at reference point $a$ by the mixed partials of the learned surrogate $\hat{g}$:
\begin{equation}
    I_S^{(r)}(a) := \left. \frac{\partial^r \hat{g}(x)}{\prod_{j \in S} \partial x_j} \right\vert{}_{x=a} , \quad r \le n.
\end{equation}

We choose a target model, which is either a transformer architecture with MHA blocks, or the ParT architecture comprising the particle attention blocks. We then train the target model to perform multiclassification on up to ten different jet flavours, taken from the \textsc{JetClass} dataset~\cite{qu_2022_6619768}.

\subsection{Extracting and Correlating Physical Alignment}
Having fit the surrogate KAN to the local decision boundary of the target model, a multi-particle interaction score can be extracted directly from its analytical gradients evaluated at the origin (the intact jet configuration). We do this by computing the second-order derivative tensor (the Hessian matrix $\mathcal{I}^{(2)}$) and the third-order derivative tensor ($\mathcal{I}^{(3)}$) of the surrogate KAN. The off-diagonal entries quantify 2-body and 3-body interactions respectively. These may additionally be interpreted as the 2nd and 3rd order Taylor Coefficients~\cite{Wunsch:2018oxb} in an expansion of the surrogate about the reference point. Thereafter, we isolate the individual contributions to $e_2^{(1)}$ $(e_3^{(1)})$ from the corresponding physical particle pairs (triplets). This physical contribution is directly compared to the extracted KAN interaction score using the Spearman rank correlation~\cite{Spearman:1904} coefficient $R_\mathrm{CORR}^{(r)}$. This acts as a metric to quantify whether substructure observables such as the ECFs align with the model's classification score. We are therefore able to gauge whether the dominant physical pairs or triplets in the ECF are also the ones that the neural network relies upon to make its decision.

\section{Results}

\begin{figure}[htbp]
  \centering
  \begin{subfigure}{0.49\textwidth}
  \centering
  \includegraphics[width=\linewidth]{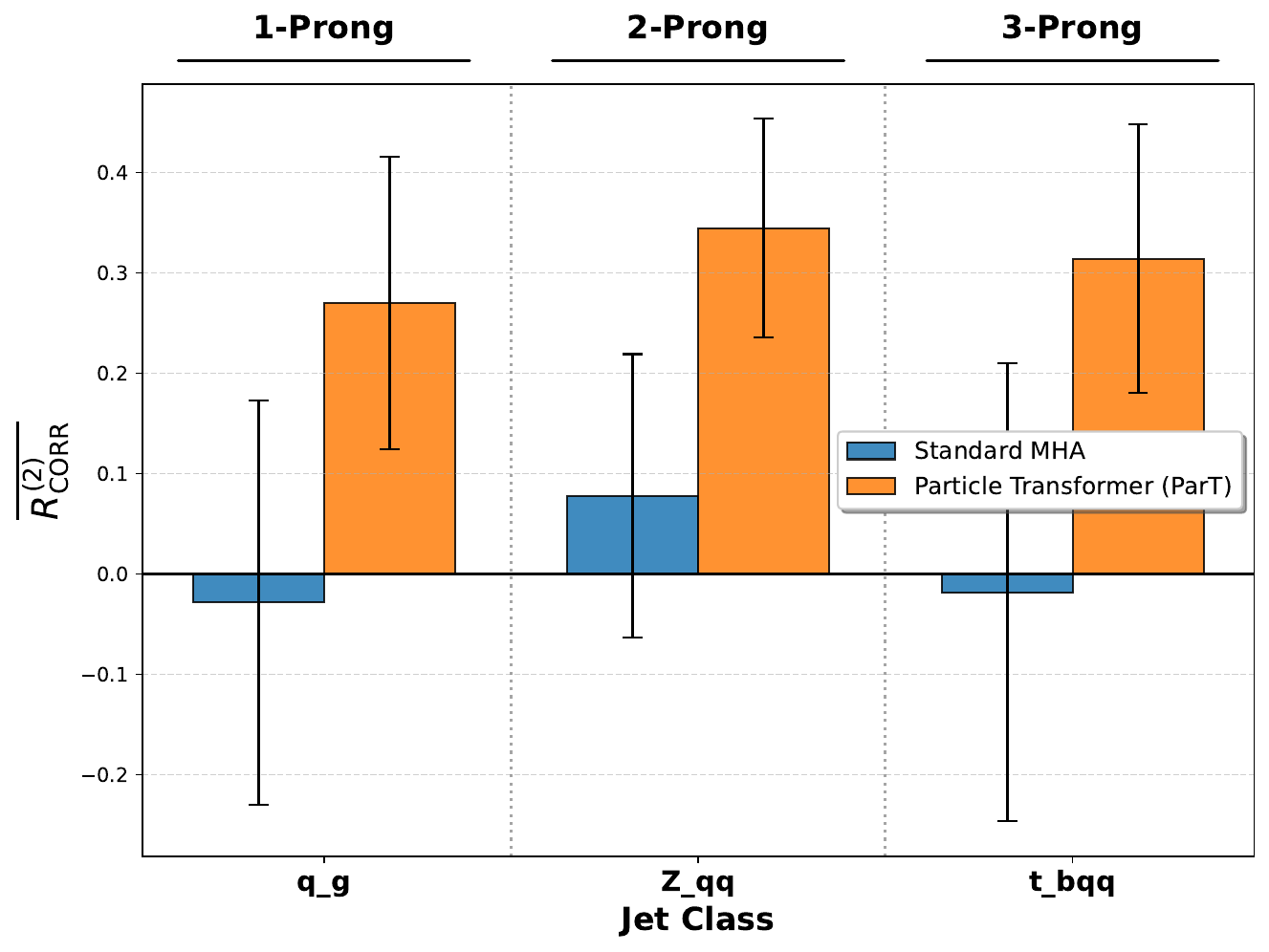}
  \caption{$\overline{R_\mathrm{CORR}^{(2)}}$ between $I_S^{(2)}$ and $e_2^{(1)}$}
    \label{fig:sink1}
  \end{subfigure}\hfill
  \begin{subfigure}{0.49\textwidth}
  \centering
  \includegraphics[width=\linewidth]{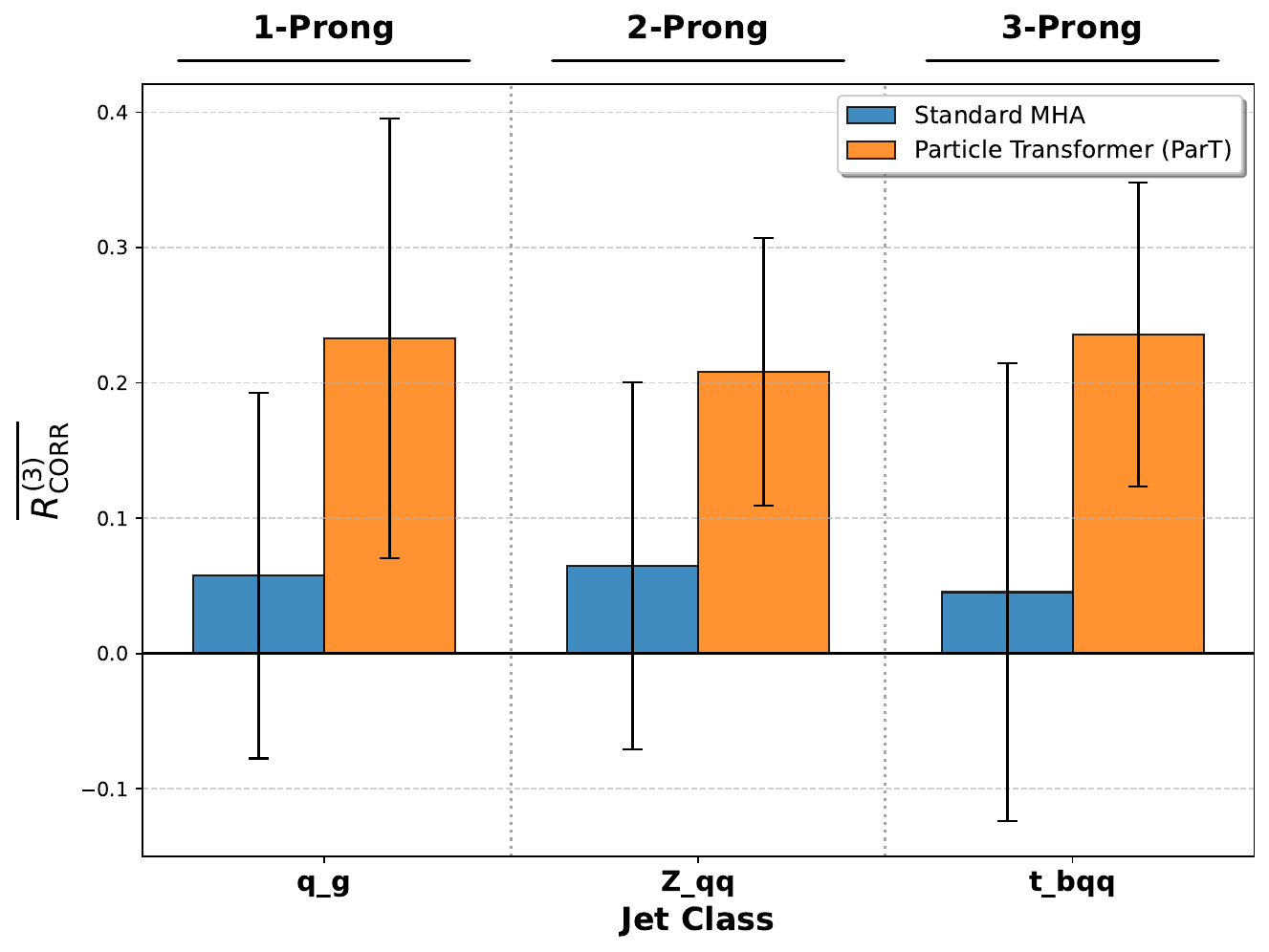}
  \caption{$\overline{R_\mathrm{CORR}^{(3)}}$ between $I_S^{(3)}$ and $e_3^{(1)}$}
    \label{fig:sink2}

  \end{subfigure}\hfill

  \caption{Ranked correlation coefficients for three representative types of jets with different radiation patterns (pronginess) show a positive correlation when using the ParT architecture for the pairwise scenario (left) for two and three-pronged jets. The triplet scenario (right) shows a weaker but still positive correlation for the same architecture.  Each bar is the mean over the 50 jets of that class for the single 8-block network of each architecture with the highest test accuracy, and the error bars show $\pm 1$ standard deviation of the per-jet correlations.}
  \label{fig:r_corr}
\end{figure}

For the purpose of this study, we compare the top 256 pairs (triplets) with the highest contribution to the energy correlators $e_2^{(1)}$ $(e_3^{(1)})$, to their extracted 2$^\mathrm{nd}$ (3$^\mathrm{rd}$)-order gradient tensors. The results are shown in Figure \ref{fig:r_corr} and an interpretation follows. 

For each jet, the surrogate is a b-spline KAN of shape $n \rightarrow 16 \rightarrow 8 \rightarrow 1$ (with $n \leq 32$ the number of constituents), using cubic splines on 5 grid intervals plus a linear residual branch on every edge ($144n + 1224$ parameters, with at most 5832). Using a \textsc{Adam} optimizer, we fit this to 50000 perturbations of the jet with a smoothness penalty on the spline coefficients, selected among three configurations on 5000 validation perturbations. This is then evaluated on 10000 held-out perturbations.


For the two‑prong case, we report a clear positive correlation for the Particle Transformer ($R^{(2)}_{\mathrm{CORR}} \approx 0.3$) between the extracted $I_S^{(2)}$ and the corresponding ECF $e_2^{(1)}$ contribution. The standard MHA, which we treat as our baseline, does not exhibit similar alignment ($\approx 0$). We therefore interpret this as the physics‑informed pairwise bias of the Particle Transformer architecture aligning its two‑body interactions towards the energy correlator contributions. 

On the other hand for the three‑prong jets, the score $R^{(3)}_{\mathrm{CORR}}$ between $I_S^{(3)}$ and $e_3^{(1)}$ is weaker for both the baseline and the ParT model. The latter is still observed to exhibit the stronger correlation, with the error bars excluding the case of null correlation. At third order, the Particle Transformer does not receive any physics-informed inductive bias to model these irreducible interactions. We therefore interpret it as the particle attention block having a higher ability than standard MHA at identifying the highest contributing triplets. More work remains to be done to identify possible structure and alignment at orders greater than 3.  

We additionally note that while the inputs to ParT include the $\Delta R_\mathrm{ij}$ terms necessary for building the n-point ECF, the presence of such a contributory term alone does not guarantee that a model will align its output with the corresponding physical observable at arbitrary order.   

\begin{table}[htbp]
  \centering
  
  \label{tab:tagging_accuracy}
  \begin{tabular}{c cc cc}
    \toprule
     & \multicolumn{2}{c}{MHA} & \multicolumn{2}{c}{ParT} \\
    \cmidrule(lr){2-3} \cmidrule(lr){4-5}
    Blocks & Best & Mean $\pm$ std & Best & Mean $\pm$ std \\
    \midrule
    1 & 54.2 & $53.1 \pm 1.2$ & 55.3 & $52.7 \pm 1.9$ \\
    2 & 72.8 & $69.8 \pm 3.5$ & 80.5 & $78.8 \pm 1.0$ \\
    4 & 77.7 & $76.1 \pm 2.5$ & 82.2 & $81.9 \pm 0.3$ \\
    8 & 81.9 & $77.8 \pm 4.8$ & \textbf{82.8} & $82.0 \pm 0.6$ \\
    \bottomrule
  \end{tabular}
  \vspace{0.5cm}
  \caption{Ten-class \textsc{JetClass} test accuracy (\%) of the best network (highest test accuracy out of 20 independently trained networks) at each depth. The mean and standard deviation over the 20 networks are given for reference.}
\end{table}

\begin{table}[htbp]
  \centering
  \label{tab:surrogate_r2}
  \begin{tabular}{cc ccc}
    \toprule
    Model & Blocks & $q/g$ (1-prong) & $Z\to q\bar{q}$ (2-prong) & $t\to bq\bar{q}'$ (3-prong) \\
    \midrule
    \multirow{4}{*}{MHA}
      & 1 & 1.000 (0.999) & 1.000 (0.999) & 1.000 (0.998) \\
      & 2 & 0.999 (0.997) & 0.998 (0.996) & 0.998 (0.996) \\
      & 4 & 0.927 (0.893) & 0.866 (0.842) & 0.908 (0.878) \\
      & 8 & 0.875 (0.776) & 0.852 (0.725) & 0.878 (0.749) \\
    \midrule
    \multirow{4}{*}{ParT}
      & 1 & 1.000 (0.999) & 1.000 (0.999) & 0.999 (0.998) \\
      & 2 & 0.995 (0.970) & 0.999 (0.998) & 0.995 (0.968) \\
      & 4 & 0.998 (0.977) & 0.998 (0.995) & 0.996 (0.983) \\
      & 8 & 0.995 (0.980) & 0.990 (0.971) & 0.993 (0.981) \\
    \bottomrule
  \end{tabular}
  \vspace{0.5cm}
  \caption{Fidelity of the KAN surrogate: coefficient of determination $R^2$ on 10000 held-out test perturbations, given as median (mean) over 1000 jets per class (50 jets from each of 20 networks) at each depth. ParT's median of 0.99 means the KAN captures 99\% of the variance in how the network's output moves under small perturbations. MHA at 8 blocks, with medians around 0.85 and a mean around 0.75, leaves a noticeable part of the response unexplained.}
  
\end{table}





\begin{figure}[htbp]
  \centering
  \begin{subfigure}{0.49\textwidth}
    \centering
    \includegraphics[width=\linewidth]{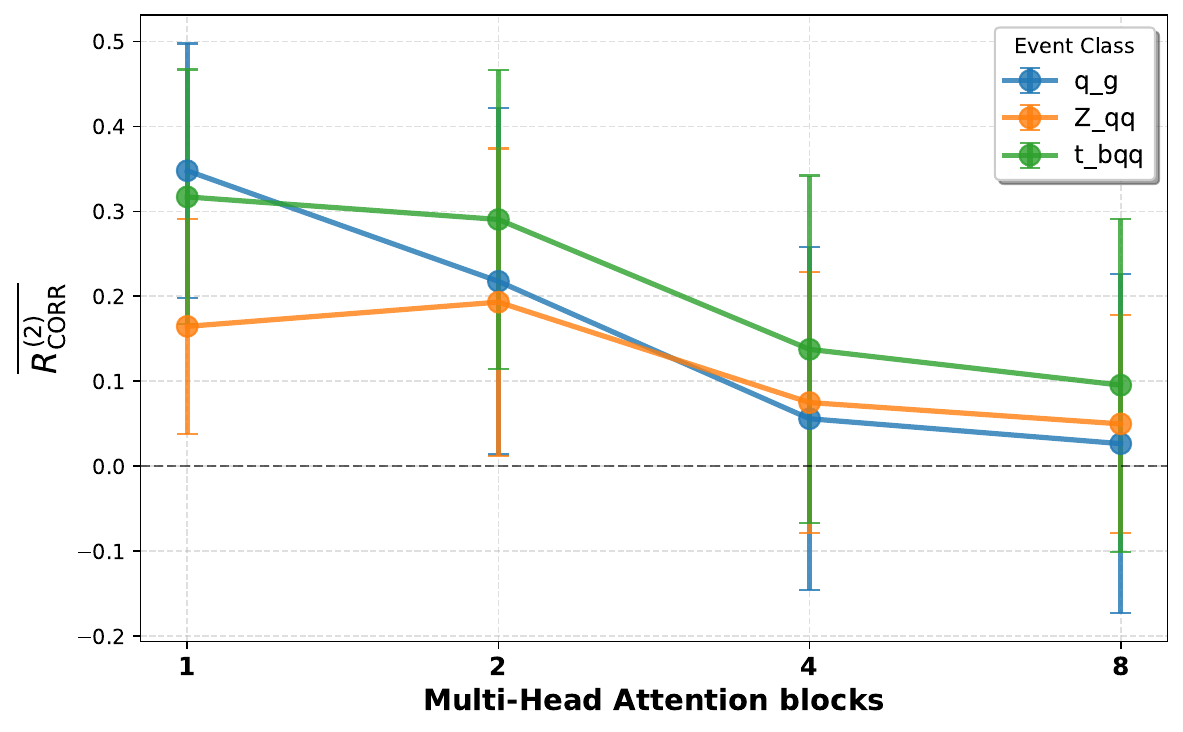}
    \caption{$\overline{R_\mathrm{CORR}^{(2)}}$ between $I_S^{(2)}$ and $e_2^{(1)}$ for MHA}
    \label{fig:mha_e2}
  \end{subfigure}\hfill
  \begin{subfigure}{0.49\textwidth}
    \centering
    \includegraphics[width=\linewidth]{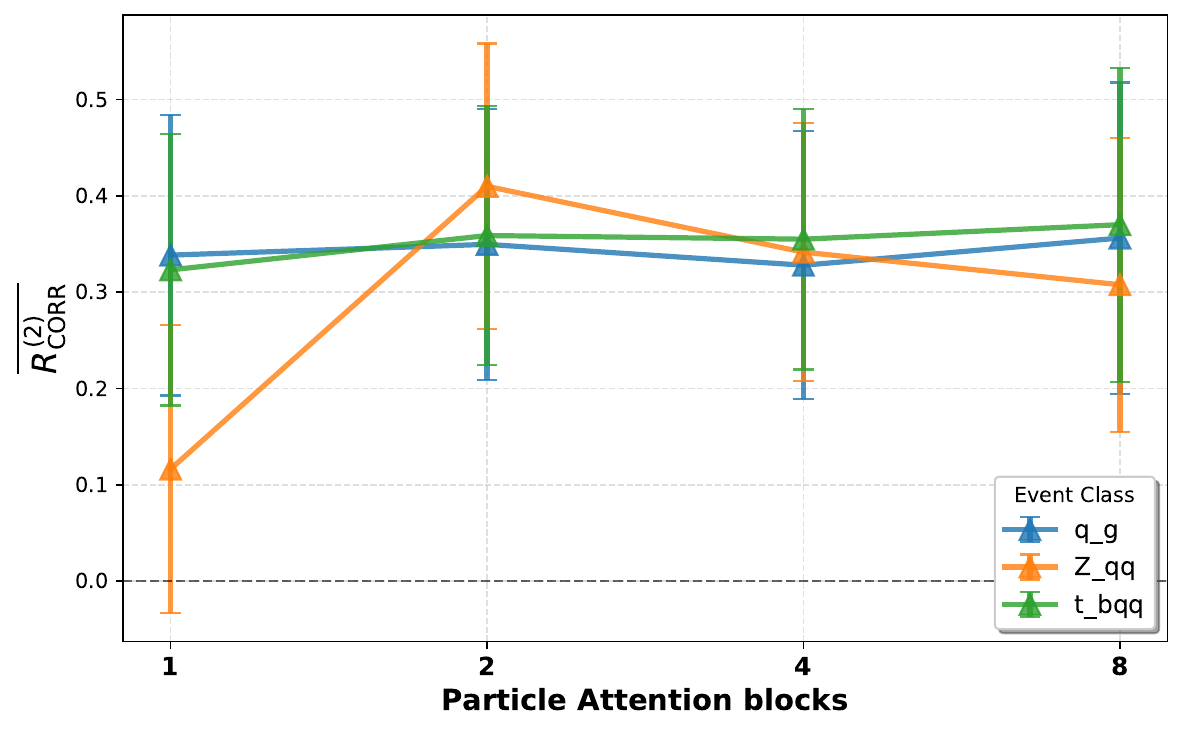}
    \caption{$\overline{R_\mathrm{CORR}^{(2)}}$ between $I_S^{(2)}$ and $e_2^{(1)}$ for ParT}
    \label{fig:part_e2}
  \end{subfigure}\hfill
  \begin{subfigure}{0.49\textwidth}
    \centering
    \includegraphics[width=\linewidth]{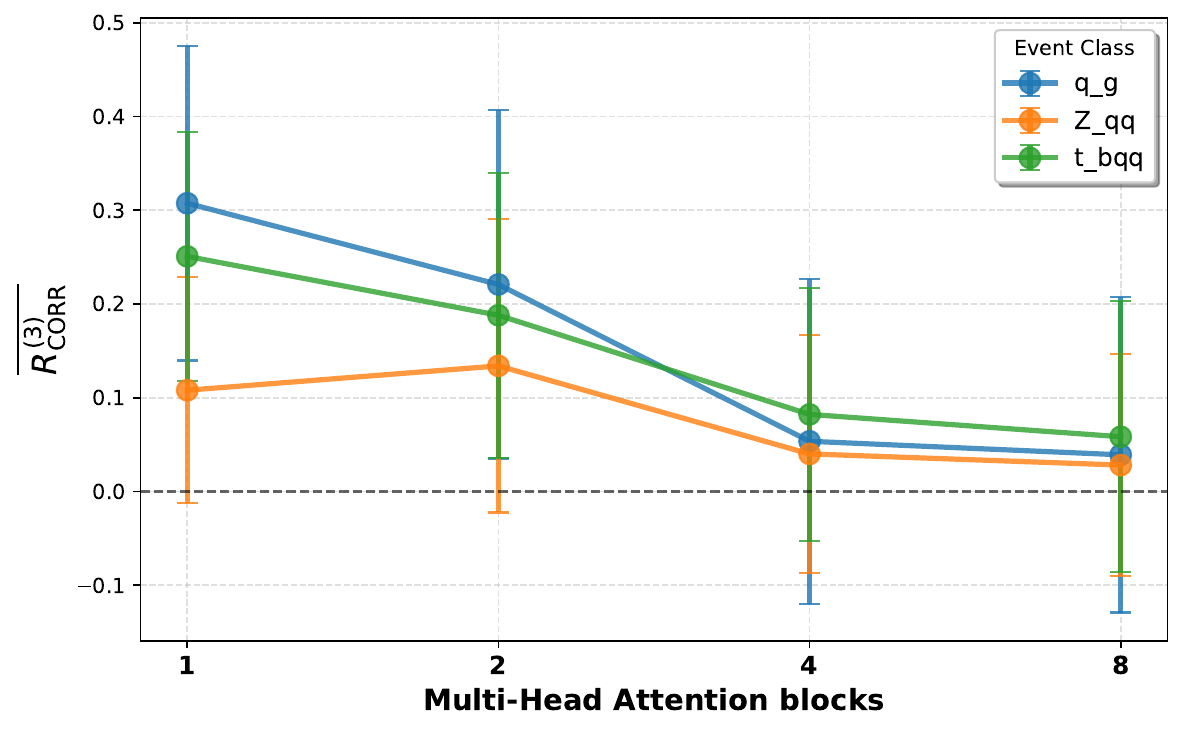}
    \caption{$\overline{R_\mathrm{CORR}^{(3)}}$ between $I_S^{(3)}$ and $e_3^{(1)}$ for MHA}
    \label{fig:mha_e3}
  \end{subfigure}
  \begin{subfigure}{0.49\textwidth}
    \centering
    \includegraphics[width=\linewidth]{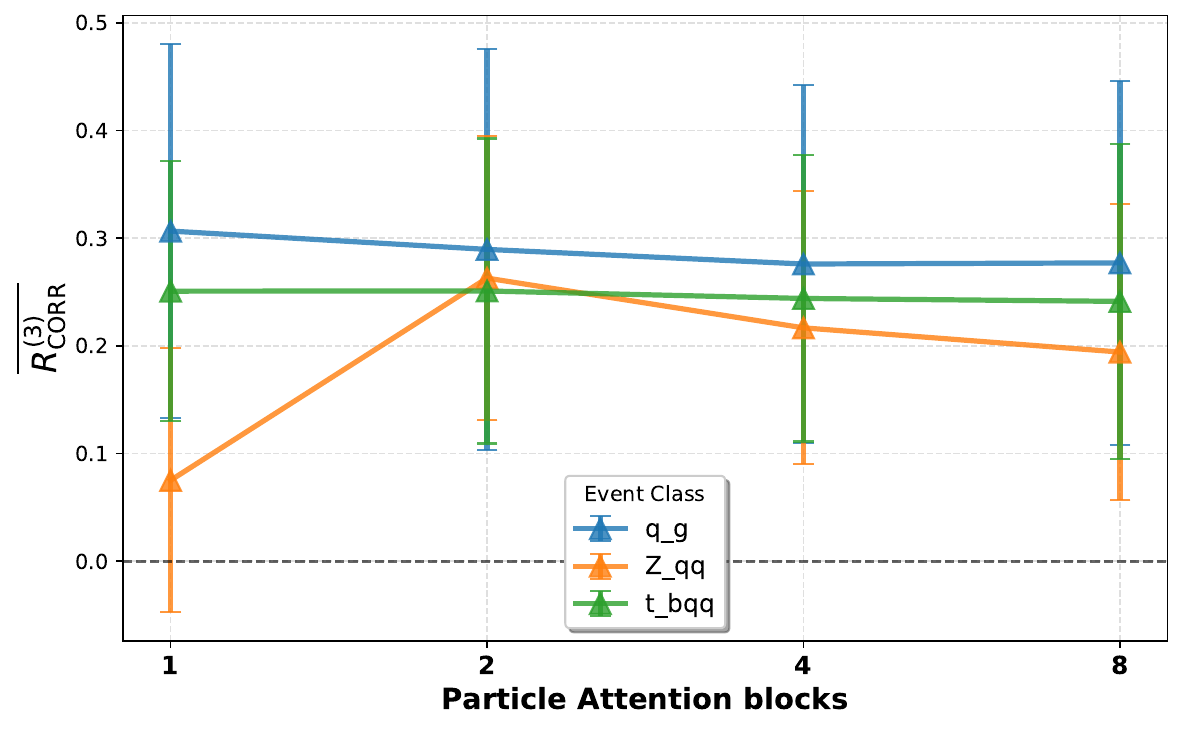}
    \caption{$\overline{R_\mathrm{CORR}^{(3)}}$ between $I_S^{(3)}$ and $e_3^{(1)}$ for ParT}
    \label{fig:part_e3}
  \end{subfigure}
  
  \caption{Rank correlation coefficients as a function of the number of transformer blocks. Additional depth is observed to have no meaningful effect on the ability to identify the necessary pairs (triplets) that dominate the ECFs at 2$^\mathrm{nd}$ (3$^\mathrm{rd}$) order. Markers show the mean over 1000 jets per class (50 jets from each of 20 independently trained networks), and error bars show $\pm 1$ standard deviation of the per-jet correlations.}
  \label{fig:depth_vs_corr}
\end{figure}

In Figure~\ref{fig:depth_vs_corr}, we examine the effect of adding subsequent MHA or Particle Attention blocks. Existing work~\cite{bal2026informationgeometryjetsubstructure} provides an interpretation that ties together higher order cumulant tensors, energy correlators and hypergraphs. It would be expected that it is the transformer depth that helps the network learn higher-order irreducible structure through consecutive layers of pair-wise attention, as opposed to requiring specialized architectures such as hypergraph neural networks. However, we report that the effect for MHA is negative and adding further blocks leads to lower alignment. For the case of the more complex ParT, this stays consistent even after adding more attention blocks. for the specific case of the 2-prong decay of the $Z$ boson, we observe anomalous behavior for both the 2- and 3-point correlations. We therefore conclude that adding more attention blocks does not meaningfully influence the ability of the model to learn higher-order structure, which would motivate the usage of more specific physics-informed architectures.

\FloatBarrier

\section{Conclusion}

In this work, we show that higher-order component attribution via a learned surrogate KAN allows us to directly map physical descriptors (such as the ECFs) of a jet's internal radiation patterns to the decision boundary of transformer-based networks. We show that for a state-of-the-art architecture such as ParT, there exists a clear correlation between the pairs (triplets) that the network identifies as important via gradient-based ranking, and the corresponding physical pairs (triplets) that contribute dominantly to the two(three)-point ECF. This is especially true for the classes of signal models (with two- and three-pronged substructure) that we examine in this work. We also demonstrate that additional model capacity via more attention blocks does not automatically translate to improved pair (triplet) identification. Further work remains to be done to extend this methodology to larger datasets, a wider class of signal models with more exotic substructure, and to higher orders beyond 3.   


\section{Responsible-use Statement}

This work maps together a novel interpretability method and a physical descriptor of energy flow patterns within a jet. While we limited the interpretability to transformer models only, this method allows for the broader community to directly probe the learning outcome of their architectures. 

\section*{Acknowledgments}

Work done at Imperial College is funded by the Science and Technology Facilities Council (STFC) grant ST/W000636/1. Work done at KIT is supported by the Alexander von Humboldt Foundation and the Helmholtz Association of German Research Centres.

\bibliography{references}    

\clearpage

\end{document}